\documentclass[conference]{IEEEtran}
\IEEEoverridecommandlockouts
\usepackage{cite}
\usepackage{amsmath,amssymb,amsfonts}
\usepackage{graphicx}
\usepackage{textcomp}
\usepackage{xcolor}
\def\BibTeX{{\rm B\kern-.05em{\sc i\kern-.025em b}\kern-.08em
    T\kern-.1667em\lower.7ex\hbox{E}\kern-.125emX}}
\usepackage[nolist]{acronym} 
\usepackage{url}            
\usepackage{booktabs}       
\usepackage{circledsteps}
\usepackage{pgfplots}
\usepackage{tikz}
\usepackage{algorithm}
\usepackage{algpseudocode}
\algdef{SE}[DOWHILE]{Do}{EndDoWhile}{\algorithmicdo}[1]{\algorithmicwhile\ #1}
\usepackage[caption=false,font=footnotesize]{subfig}
\usepackage{multirow}
\usepackage[table]{xcolor}
\pgfplotsset{compat=1.18}
\definecolor{algblue}{RGB}{31,78,121}
\definecolor{alggreen}{RGB}{0,117,94}
\definecolor{algorange}{RGB}{190,96,0}
\newcommand{\AlgPhase}[2]{\Statex \textcolor{#1}{\textbf{\(\triangleright\) #2}}}
\usepackage{xurl}
\usepackage[hidelinks]{hyperref}
\begin{document}

\title{Exp2VLA: Enabling Vision–Language–Action \\ for Drone Navigation from Expert Demonstrations
\thanks{*This work is supported in part by the Horizon Europe Grant Agreements No.~101136056 and No.~101119774.}
\thanks{$\ddagger$These authors contributed equally to this work.}
}
\author{\IEEEauthorblockN{1\textsuperscript{st} Van Huyen Dang$^{\ddagger}$}
\IEEEauthorblockA{\textit{Automatic Control Group}\\ \textit{Paderborn University}\\
Paderborn, Germany \\
van.huyen.dang@upb.de}
\and
\IEEEauthorblockN{2\textsuperscript{nd} Kabilesh Rajendran$^{\ddagger}$}
\IEEEauthorblockA{\textit{Automatic Control Group}\\ \textit{Paderborn University}\\
Paderborn, Germany \\
kabilesh@mail.uni-paderborn.de}
\and
\IEEEauthorblockN{3\textsuperscript{rd} Erdi Sayar}
\IEEEauthorblockA{\textit{Automatic Control Group}\\ \textit{Paderborn University}\\
Paderborn, Germany \\
erdi.sayar@upb.de}
\and
\IEEEauthorblockN{4\textsuperscript{th} Erdal Kayacan}
\IEEEauthorblockA{\textit{Automatic Control Group}\\ \textit{Paderborn University}\\
Paderborn, Germany \\
erdal.kayacan@upb.de}
}

\maketitle

\begin{acronym}
\acro{rl}[RL]{reinforcement learning}
\acro{vla}[VLA]{vision-language-action}
\acro{uav}[UAV]{unmanned aerial vehicle}
\acro{Exp2VLA}[Exp2VLA]{expert-to-VLA pipeline}

\end{acronym}

\begin{abstract}
Vision-language-action (VLA) models open a new path toward intuitive robot control by directly linking perception, language, and action in a single end-to-end framework. Yet for UAVs, practical adoption remains difficult because existing solutions are either computationally heavy or insufficiently capable in complex environments. In this work, we propose a practical expert-distillation pipeline (Exp2VLA) for language-conditioned drone navigation. The core idea is to distill expert behavior, obtained from reinforcement learning, teleoperation, or other controllers, into training data that can be used to fine-tune compact VLA models. This allows existing control strategies to be transferred into a unified language-guided navigation model, reducing manual system integration and lowering the barrier for deploying new robot behaviors. Experiments in both sim-to-sim and simulation-in-the-loop settings across multi-object scenes show that the fine-tuned models can handle varied semantic commands and generalize to unseen target compositions. The proposed framework demonstrates how expert-policy distillation can help mechatronic systems move from specialized control modules toward more flexible and reusable robot intelligence.
\end{abstract}


\section{Introduction}\label{sec:introduction}
\IEEEPARstart{R}{obotic} systems, including aerial robots,  operate under diverse physical constraints and environments. Consequently, different platforms require different strategies for perception, decision-making, and control, as well as different mechatronic designs that tightly couple sensing, actuation, and computation. 

\begin{figure}[t!]
    \centering
    \includegraphics[width=0.92\linewidth, trim=0cm 0cm 0 0,clip]{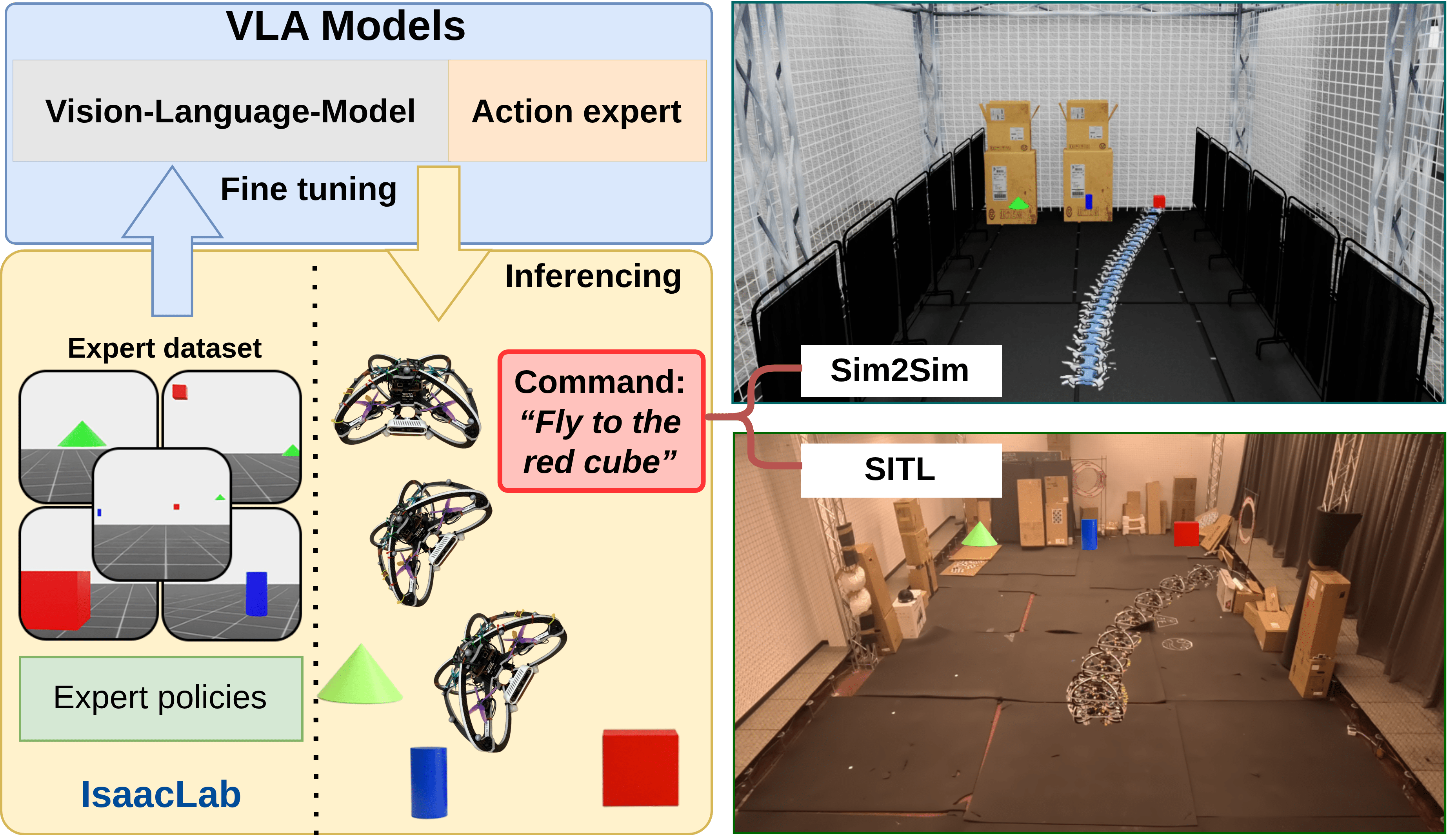}
    \caption{Overview of the proposed Exp2VLA pipeline. Expert policies in Isaac Lab generate demonstration rollouts, which are converted into a dataset for fine-tuning a vision-language-action model. At inference time, the fine-tuned model takes RGB observations and language commands as input and produces control actions for drone navigation. We evaluate the resulting policy in both sim-to-sim and simulation-in-the-loop (SITL) settings.
}
    \label{fig:frontimage}
\end{figure}

Recent advances in deep learning offer an alternative paradigm in which robots learn complex behaviors directly from data~\cite{pham2022deep, pham2022pencilnet, dang2025vdsnav}. Learning-based approaches can integrate perception and control within unified architectures, reducing reliance on hand-crafted intermediate representations. More recently, \ac{vla} models (see Fig.~\ref{fig:frontimage}) have emerged as a promising direction toward more general and flexible robotic systems. These models jointly process visual observations and language instructions to produce robot actions, enabling robots to reason about tasks, environments, and their underlying mechatronic dynamics in a more expressive manner.
Building on the potential of \ac{vla} models for complex real-world tasks, we present \acs{Exp2VLA}, an expert-distillation pipeline for language-conditioned drone navigation. Our approach distills expert behavior into training data for fine tuning a compact \ac{vla} model, enabling existing control strategies to be transferred into an end-to-end aerial navigation model. In this way, \ac{Exp2VLA} moves \ac{vla}-based drone control closer to practical deployment on resource-constrained platforms.

We summarize our contributions as follows:
\begin{itemize}
    \item We introduce \ac{Exp2VLA}, an expert-distillation pipeline that distills expert demonstrations from reinforcement learning, teleoperation, or other controllers into training data for \ac{vla}-based drone navigation.
    \item We integrate expert-driven data collection in simulation, automated conversion to the LeRobot format, and end-to-end policy adaptation into a unified framework for language-conditioned aerial control.
    \item We validate the proposed approach through both sim-to-sim and simulation-in-the-loop experiments, showing that fine-tuned \acp{vla} with \ac{Exp2VLA} can follow semantic commands across multi-object scenes and generalize to unseen object--color compositions.
    \item We publicly release our datasets to support reproducibility and future research\footnote{Datasets are available at \url{https://huggingface.co/datasets/UPB-RAT-VLA/Exp2VLA-SingleCube-v1} and \url{https://huggingface.co/datasets/UPB-RAT-VLA/Exp2VLA-MultiObject-v1}.}.
\end{itemize}

The paper is structured as follows: Section~\ref{sec:relatedwork} summarizes related work, while Section~\ref{sec:methodology} presents the proposed method. Section~\ref{sec:sim_study} describes the simulation setup and experimental evaluation, and Section~\ref{sec:conclusion} concludes with insights and directions for future work.
\section{Related work}\label{sec:relatedwork}
Recent works have applied \ac{vla} models to mobile and embodied robots beyond fixed-base manipulation. For instance, \cite{cheng2024navila} proposes a hierarchical framework for legged robot navigation, in which a \ac{vla} model generates mid-level language commands that are executed by a low-level locomotion policy. Similarly, \cite{huang2025mobilevla} introduces a unified \ac{vla} framework for quadruped robots that grounds natural-language instructions into continuous control actions, supported by the MobileVLA-CoT dataset and a two-stage training strategy combining supervised chain-of-thought learning with reinforcement learning. In \cite{wu2025momanipvla}, pre-trained \ac{vla} models are adapted from fixed-base manipulators to mobile manipulators using \ac{vla}-generated waypoints. Taken together, these studies show that \ac{vla} models are increasingly being extended from manipulation toward broader embodied control settings.

This trend naturally motivates the study of \ac{vla}-based control for aerial robots. However, transferring these advances to \acp{uav} is nontrivial because aerial robotics lacks the kind of large-scale, domain-specific demonstration data that has supported progress in manipulation and ground robotics. In particular, datasets capturing expert behavior from first-person flight perspectives remain scarce. As a result, effectively training or adapting \ac{vla} models for autonomous \ac{uav} operation is substantially more difficult.

Despite these challenges, recent studies have begun to explore \ac{vla}-based aerial robotics, although most current approaches rely on large-scale architectures such as OpenVLA-7B~\cite{openvla}. For example, CognitiveDrone~\cite{cognitivedrone2024} introduces a dedicated \ac{vla} framework for \acp{uav} that maps first-person visual observations and language instructions directly to continuous control commands through explicit vision--language reasoning modules. AutoFly~\cite{sun2025autofly} further extends this direction to autonomous outdoor navigation with a two-stage training strategy and reports both simulation and real-world results. Likewise, RaceVLA~\cite{racevla2025}, one of the first \ac{vla} systems designed for high-speed autonomous drones, generates linear and angular velocity commands from first-person-view imagery and language input, but still requires OpenVLA-scale models running on a stationary computer via a network interface. These works establish the feasibility of language-conditioned aerial control, but they also highlight the practical cost of relying on large models.

A closely related line of research can be found in recent \ac{uav} vision-language-navigation (VLN) systems, which likewise push toward language-grounded aerial autonomy. In \ac{uav}-VLN~\cite{uavvln2025}, end-to-end language-guided \ac{uav} navigation with cross-modal grounding is studied, while VLFly~\cite{vlfly2025} emphasizes open-vocabulary goal understanding for instruction-conditioned flight. The unified aerial VLN framework in~\cite{aerialvln2026} further shows that strong navigation performance can be achieved using monocular RGB observations and prompt-guided multi-task learning. While these studies demonstrate rapid progress in language-conditioned aerial navigation, they are primarily formulated as VLN systems rather than compact end-to-end \ac{vla} control policies for high-frequency closed-loop deployment.

This raises the question of what model scale is most suitable for deployable aerial VLA systems. Although recent compact VLA models aim to reduce inference cost while retaining multimodal reasoning, sub-billion-parameter architectures such as SmolVLA~\cite{smolvla} often lack sufficient capacity for complex spatial reasoning in multi-object scenes with longer temporal horizons. At the other extreme, 7B-class models deliver strong semantic reasoning but typically exceed the computational budgets of edge-deployed UAV platforms. The $\pi_{0.5}$ architecture represents a promising middle ground. Built on the 3B-parameter PaliGemma vision--language backbone with a dedicated 300M-parameter Flow Matching action expert, it supports high-frequency reactive control while preserving strong semantic reasoning and cross-embodiment generalization~\cite{pi05, paligemma, liberopro}.

Nevertheless, compact and mid-scale \ac{vla} architectures remain largely unexplored in aerial robotics. Most existing \ac{uav}-oriented \ac{vla} systems still rely on large-scale models, leaving open the question of whether smaller multimodal architectures can provide sufficient reasoning capability while remaining practical for resource-constrained \acp{uav}. Addressing this question requires understanding how multimodal reasoning scales with model size, how limited aerial demonstration data can be leveraged effectively, and how real-time closed-loop control can be maintained under strict computational constraints. These open issues motivate a systematic investigation of compact \ac{vla} architectures tailored to aerial deployment.
\section{Methodology}\label{sec:methodology}
\begin{figure*}[!t]
    \centering
    \vspace{5pt}
      \includegraphics[width=0.92\textwidth]{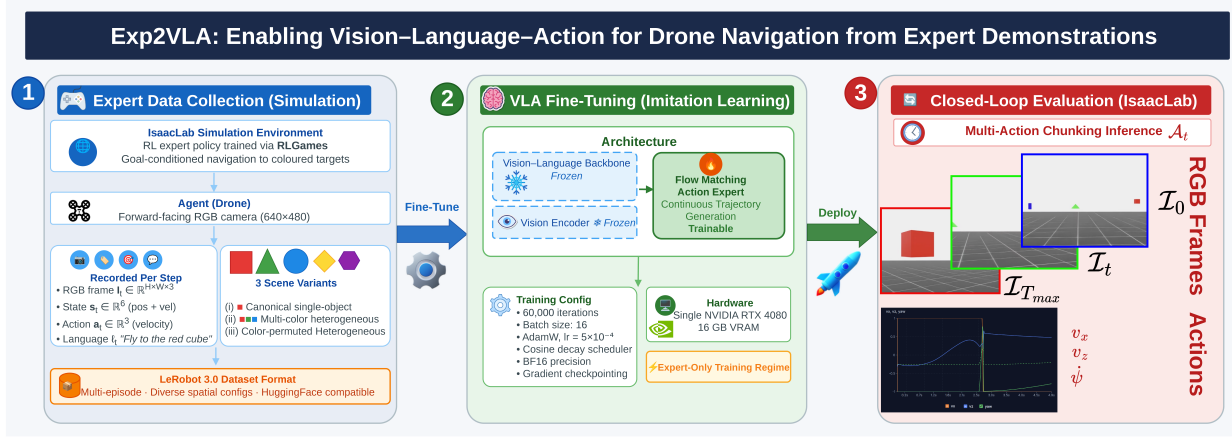}
    \caption{The \ac{Exp2VLA} pipeline consists of three stages: (1) collecting expert demonstrations in Isaac Lab and converting them into the LeRobot format, (2) fine-tuning a VLA model via imitation learning, and (3) deploying the fine-tuned models in closed loop for language-conditioned drone navigation in sim-to-sim.}
    \label{fig:pipeline}
\end{figure*}

We propose an end-to-end framework for robotics that can gather expert demonstrations from diverse sources, including manual teleoperation, optimal control strategies, and well-trained policies in simulation, and converts them into structured datasets. These datasets enable the fine-tuning of \ac{vla} models such as SmolVLA and $\pi_{0.5}$, allowing them to be adapted to a wide range of robotic tasks and platforms. The overall pipeline, illustrated in Fig.~\ref{fig:pipeline}, is referred to as \ac{Exp2VLA}.
\subsection{Expert data collection}
\label{sec:data_collection}

While the collection of expert demonstrations in our pipeline is inherently agnostic to the underlying control scheme, accommodating classical control, human teleoperation, or learning-based policies, we opt for an automated approach. Specifically, to establish a robust pre-trained expert for demonstration collection, we train a continuous-control navigation policy using the proximal policy optimization (PPO) \cite{schulman2017proximal} algorithm via the RLGames framework~\cite{rl-games2021} within Isaac Lab~\cite{mittal2025isaac}.
The resulting policy serves as an oracle, receiving explicit spatial goal coordinates to guide the agent (\ac{uav}) toward visually defined targets (e.g., colored objects). The goal is set at 0.5m offset in front of the visual target to aid in fine-tuning of the VLA models. Once trained, this goal-directed expert is executed in the simulation environment to systematically record high-quality demonstration data across multiple episodes.

At every control step, the following quantities are recorded:
\begin{itemize}
\setlength{\itemsep}{0pt}\setlength{\parsep}{0pt}\setlength{\parskip}{0pt}
    \item an RGB frame \(\mathbf{I}_t \in \mathbb{R}^{H \times W \times 3}\) (with \(H=480\), \(W=640\)) 
          from the drone's forward-facing camera,
    \item the drone state vector \(\mathbf{s}_t \in \mathbb{R}^{6}\) comprising linear and angular velocities in body frame,
    \item the continuous velocity action \(\mathbf{a}_t \in \mathbb{R}^{3}\) executed at that step, including forward ($v_x$) and vertical ($v_z$) velocities, and yaw rate ($\dot{\psi}$).
    \item a natural-language task instruction \(\ell_t\) specifying the desired behavior (e.g., \texttt{Fly to the red cube}), which is tokenized into a numerical representation using the model's native tokenizer to facilitate multimodal integration.
\end{itemize}

Data collection spans two environment variants of increasing complexity: single-object scenes and multi-color heterogeneous scenes (see Section~\ref{sec:exp_setup}) to expose the policy to diverse spatial configurations and varied combinations of object color and shape. We convert the resulting raw data into the LeRobot~\cite{capuano2025robot} dataset format to ensure compatibility with standard \ac{vla} training and fine-tuning pipelines.

\subsubsection{Dataset characteristics}
Table~\ref{tab:dataset_stats} summarizes the key properties of the curated heterogeneous scene training dataset. The dataset comprises 1{,}500 episodes with a fixed episode length of 247~steps yielding a total of 370{,}500 annotated timesteps. Each timestep includes a $640\times480$ RGB frame encoded with the AV1 codec, a six-dimensional state observation, and a three-dimensional action command, resulting in a compact dataset size of approximately 0.98\,GB. Three distinct natural-language instructions are used, corresponding to the three target objects in the heterogeneous scene environment.

Table~\ref{tab:state_action_stats} reports per-dimension statistics for the observation state and action vectors for two different data sets:
\begin{itemize}
\setlength{\itemsep}{0pt}\setlength{\parsep}{0pt}\setlength{\parskip}{0pt}
    \item Dataset~1 (Exp2VLA-SingleCube-v1) belongs to the data set with a single object (which has only red cubes inside).
    \item Dataset~2 (Exp2VLA-MultiObject-v1) includes different types of objects and different types of colors.
\end{itemize}
\vspace{-5pt}
The state statistics reveal that across both datasets, the drone predominantly moves along the forward axis ($v_x$: mean\,$\approx$\,1.23\,m/s for Dataset~1 and 1.14\,m/s for Dataset~2), with minimal lateral ($v_y \approx 0$) and vertical ($v_z \approx 0$) drift, consistent with the goal-directed nature of the navigation task. Orientation remains nearly level throughout ($\dot{\phi}$ and $\dot{\theta}$ rates\,$\approx 0$), while the yaw rate exhibits moderate variation ($\sigma \approx 0.14$\,rad for Dataset~1 and 0.15\,rad for Dataset~2) as the drone adjusts its heading toward the target. In the action space, the active dimensions ($v_x$, $v_z$, and $\dot{\psi}$) span the full normalized range $[-1, 1]$ in both datasets, indicating diverse control commands. The other potential control dimensions ($v_y$, pitch, and roll rates) remain unused, reflecting the constrained action manifold of the velocity controller. This effectively reduces the action space to $\mathbb{R}^3$, as described in Section~\ref{sec:data_collection}.

\begin{table}[b!]
  \centering
  \caption{Summary of the RGB-shapes training dataset.}
  \label{tab:dataset_stats}
  \small
  \renewcommand{\arraystretch}{0.92}
  \setlength{\tabcolsep}{5pt}
  \begin{tabular}{lr}
    \toprule
    \textbf{Property} & \textbf{Value} \\
    \midrule
    Total episodes           & 1{,}500 \\
    Total timesteps          & 370{,}500 \\
    Episode length           & 247 steps (4.94\,s) \\
    Image resolution         & $640\times480\times3$ \\
    Video codec              & AV1 \\
    Unique language instructions & 3 \\
    Dataset size (on disk)   & 0.98\,GB \\
    \bottomrule
  \end{tabular}
\end{table}

\begin{table}[b!]
  \centering
  \caption{Per-dimension statistics of observation states and action
           commands for Dataset~1 and Dataset~2.
           Values with magnitude below $10^{-6}$ are reported as~$0$.}
  \label{tab:state_action_stats}
  \renewcommand{\arraystretch}{0.93}
  \setlength{\tabcolsep}{2.5pt}
  \resizebox{0.93\columnwidth}{!}{%
  \scriptsize
  \begin{tabular}{cl rrrr rrrr}
    \toprule
    & & \multicolumn{4}{c}{\textbf{Dataset~1}$^{*}$}
      & \multicolumn{4}{c}{\textbf{Dataset~2}$^{\dagger}$} \\
    \cmidrule(lr){3-6} \cmidrule(lr){7-10}
    \textbf{Dim} & \textbf{Name}
      & \textbf{Mean} & \textbf{Std} & \textbf{Min} & \textbf{Max}
      & \textbf{Mean} & \textbf{Std} & \textbf{Min} & \textbf{Max} \\
    \midrule
    \multicolumn{10}{l}{\textit{Observation state}} \\
    $0$ & $v_x$          & $1.2340$ & $0.9560$ & $0$        & $2.0000$
                          & $1.1354$ & $0.9704$ & $0$        & $2.0000$ \\
    $1$ & $v_y$          & $0.0005$ & $0.0054$ & $-0.0105$  & $0.0105$
                          & $0.0024$ & $0.0057$ & $-0.0105$  & $0.0105$ \\
    $2$ & $v_z$          & $0.0476$ & $0.1077$ & $-0.5000$  & $0.5000$
                          & $-0.0006$ & $0.1361$ & $-0.5000$ & $0.5000$ \\
    $3$ & $\dot{\phi}$   & $0$      & $0$      & $0$        & $0$
                          & $0$      & $0$      & $0$        & $0$      \\
    $4$ & $\dot{\theta}$ & $0$      & $0$      & $0$        & $0$
                          & $0$      & $0$      & $0$        & $0$      \\
    $5$ & $\dot{\psi}$   & $0.0180$ & $0.1446$ & $-0.2617$  & $0.2617$
                          & $0.0253$ & $0.1540$ & $-0.2617$  & $0.2617$ \\
    \midrule
    \multicolumn{10}{l}{\textit{Action command}} \\
    $0$ & $v_x$          & $0.2481$ & $0.9666$ & $-1.0000$  & $1.0000$
                          & $0.1550$ & $0.9860$ & $-1.0000$  & $1.0000$ \\
    $1$ & $v_z$          & $0.2736$ & $0.2573$ & $-1.0000$  & $1.0000$
                          & $0.2020$ & $0.3610$ & $-1.0000$  & $1.0000$ \\
    $2$ & $\dot{\psi}$   & $0.0687$ & $0.5527$ & $-1.0000$  & $1.0000$
                          & $0.0970$ & $0.5890$ & $-1.0000$  & $1.0000$ \\
    \bottomrule
  \end{tabular}%
  \normalsize
  }\\[0pt]
  \raggedright\scriptsize
  An interactive visualization of these datasets is publicly available:\\
  $^{*}$\;\href{https://huggingface.co/spaces/lerobot/visualize_dataset?path=%2FUPB-RAT-VLA%2FExp2VLA-SingleCube-v1%2Fepisode_0}{huggingface.co/Exp2VLA-SingleCube-v1}\\
  $^{\dagger}$\;\href{https://huggingface.co/spaces/lerobot/visualize_dataset?path=%2FUPB-RAT-VLA%2FExp2VLA-MultiObject-v1%2Fepisode_0}{huggingface.co/Exp2VLA-MultiObject-v1}
\end{table}
\subsection{Fine-tuning VLAs with Exp2VLA}
\label{sec:train_vla}
\begin{figure}[!t]
\centering

\subfloat[Training loss.]{
\includegraphics[width=0.42\linewidth,
trim=0cm 0cm 0 0,clip]{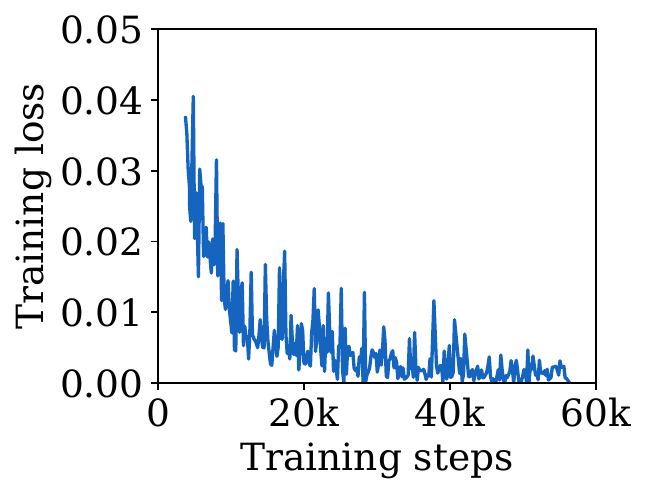}
\label{fig:loss_curve}}
\hfill
\subfloat[Gradient norm.]{
\includegraphics[width=0.42\linewidth,
trim=0 0cm 0 0,clip]{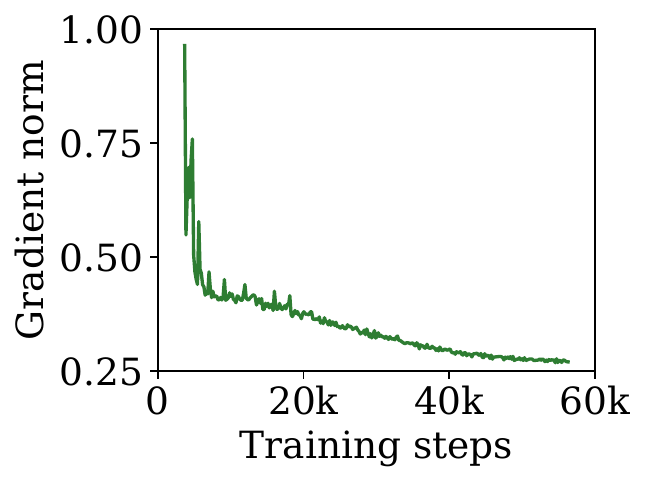}
\label{fig:grad_norm_curve}}

\caption{Training metrics for fine-tuning $\pi_{0.5}$ on datasets obtained via \ac{Exp2VLA} over $60k$ steps (batch size 16, $\text{lr}=5e^{-4}$, BF16 precision). The SigLIP vision encoder is frozen and only the Flow Matching action expert is updated.}
\label{fig:wbloss}

\end{figure}

We fine-tune existing \ac{vla} models on our curated dataset via imitation learning, focusing on the 4B-parameter $\pi_{0.5}$ and the ultra-compact 0.45B-parameter SmolVLA~\cite{smolvla}. To maintain high throughput on a single consumer-grade NVIDIA GeForce RTX 4080 Super (16\,GB), we use bfloat16 (BF16) numerical precision for model weights and activations, ensuring a stable dynamic range while significantly reducing the VRAM footprint. To further fit the architectures into limited device memory, we enable gradient checkpointing and freeze the vision encoders (e.g., the 400M-parameter SigLIP~\cite{zhai2023siglip} for $\pi_{0.5}$), focusing updates exclusively on the action prediction modules. This ``expert-only'' training regime allows the models to specialize in robotic control while preserving the general-purpose semantic knowledge of their frozen backbones. The policies are optimized for $60{,}000$ iterations with a batch size of 16 for $\pi_{0.5}$ and 64 for SmolVLA, and the learning rate is set to $5 \times 10^{-4}$ with a cosine decay scheduler based on the total step count. The training loss as well as the gradient norm for the $\pi_{0.5}$ fine-tuning is shown in Fig.~\ref{fig:wbloss}.

\subsection{Closed-loop evaluation in simulation}
The fine-tuned $\pi_{0.5}$ policy is deployed back into Isaac Lab for closed-loop evaluation, following the procedure detailed in Algorithm~\ref{alg:vla_drone_inference}. 
\begin{algorithm}[t!]
\caption{Closed-loop \ac{vla} inference for drone navigation}
\label{alg:vla_drone_inference}
\begin{algorithmic}[1]
\scriptsize
\Require Policy $\pi_\theta$, environment $\mathcal{E}$, command $\ell$, target $\mathbf{g}$, chunk size $K$, horizon $T_{\max}$, radius $\epsilon$
\Ensure \textsc{Success} if the drone reaches $\mathbf{g}$ within $\epsilon$; otherwise \textsc{Failure}

\AlgPhase{algblue}{Initialize episode and language context}
\State $\mathbf{s}_0,\,\mathbf{I}_0,\,\mathbf{p}_0 \gets \mathcal{E}.\textsc{Reset}()$
  \Comment{State, image, position}
\State $\mathbf{z}_{\mathcal{L}} \gets \textsc{Tokenize}(\ell)$
  \Comment{Fixed command tokens}
\State $t \gets 0$

\While{$t < T_{\max}$}

    \AlgPhase{alggreen}{Build observation and infer an action chunk}
    \State $\hat{\mathbf{s}}_t \gets \textsc{Normalize}(\mathbf{s}_t)$
      \Comment{Use training statistics}

    \State $\mathbf{o}_t \gets (\mathbf{I}_t,\;\hat{\mathbf{s}}_t,\;\mathbf{z}_{\mathcal{L}})$
      \Comment{Multimodal observation}

    \State $\mathbf{A}_{t:t+K-1} \in \mathbb{R}^{K \times 3} \gets \pi_\theta(\mathbf{o}_t)$
      \Comment{$K$ actions: $v_x,v_z,\dot{\psi}$}

    \AlgPhase{algorange}{Execute commands}
    \For{$k = 0, 1, \ldots, K-1$}

        \State $\mathbf{u}_t \gets f_{\mathrm{vel}}(\mathbf{A}_{t:t+K-1}[k])$
          \Comment{Controller command}

        \State $\mathbf{s}_{t+1},\,\mathbf{I}_{t+1},\,\mathbf{p}_{t+1} \gets \mathcal{E}.\textsc{Step}(\mathbf{u}_t)$

        \State $t \gets t + 1$

        \If{$\lVert \mathbf{p}_{t} - \mathbf{g} \rVert_2 \leq \epsilon$}
          \State \Return \textsc{Success}
        \EndIf

        \If{$t \geq T_{\max}$}
          \State \textbf{break}
        \EndIf

    \EndFor

\EndWhile

\State \Return \textsc{Failure}

\end{algorithmic}
\end{algorithm}
In each trial, the drone is initialized at a randomized attitude, while target objects are sampled from the position intervals defined in Section~\ref{sec:exp_setup}. The model operates exclusively in inference mode: the language instruction is tokenized once at the beginning of the episode, and each inference call receives the latest first-person image, normalized drone state, and fixed language tokens.

To ensure real-time performance and reduce the computational cost of continuous \ac{vla} inference, our pipeline utilizes multi-action chunking. Instead of re-inferring at every control timestep, the policy predicts $\mathbf{A}_{t:t+K-1} \in \mathbb{R}^{K \times 3}$, a short sequence of $K$ velocity-style actions corresponding to forward velocity, vertical velocity, and yaw rate. These actions are executed sequentially by the drone controller until the chunk is consumed, the maximum horizon is reached, or the evaluator detects success. An evaluation trial is considered successful if the drone reaches and stabilizes within a predefined distance threshold $\epsilon$ of the target object. Formally, the success condition $\mathcal{S}$ is defined as:
\begin{equation}
    \mathcal{S} = \mathbf{1}\!\left[ \min_t \lVert \mathbf{p}_t - \mathbf{g} \rVert_{2} \leq \epsilon \right],
    \label{eq:success_rate}
\end{equation}
where $\mathbf{p}_t \in \mathbb{R}^3$ and $\mathbf{g} \in \mathbb{R}^3$ denote the drone's position and target coordinates, respectively. Performance is measured by the task success rate across the diverse language-conditioned navigation scenarios.

\section{Experiments}\label{sec:sim_study}

\subsection{Experimental setup}
\label{sec:exp_setup}
All experiments are conducted in Isaac Lab using a quadcopter equipped with a forward-facing RGB camera ($640\times480$). The drone operates in a continuous velocity action space. For the evaluated \ac{vla} policies, specifically the $\pi_{0.5}$ and SmolVLA backbones, the action chunk size is set to $K=50$. At the start of each episode, both the drone and the target are randomized in position to encourage robustness and prevent memorization.

The evaluation success threshold is defined as $\epsilon = 0.60$\,m. Specifically, the target object is placed at a random world-frame position with $x \in [5,7]$\,m, $y \in [-2,2]$\,m, and $z \in [0.75,1.5]$\,m relative to the environment origin. The drone is initialized with a randomized altitude $z \in [0.75,1.5]$\,m, while the horizontal position is reset to the environment origin. Each episode lasts 5 seconds, and success is defined by reaching and stabilizing within the acceptance radius $\epsilon$ of the commanded object (Eq.\ref{eq:success_rate}).

\begin{figure}[!t]
\centering

\subfloat[Canonical single-object scene (Dataset~1).]{
\includegraphics[width=0.3\linewidth,
trim=0cm 0cm 0 0,clip]{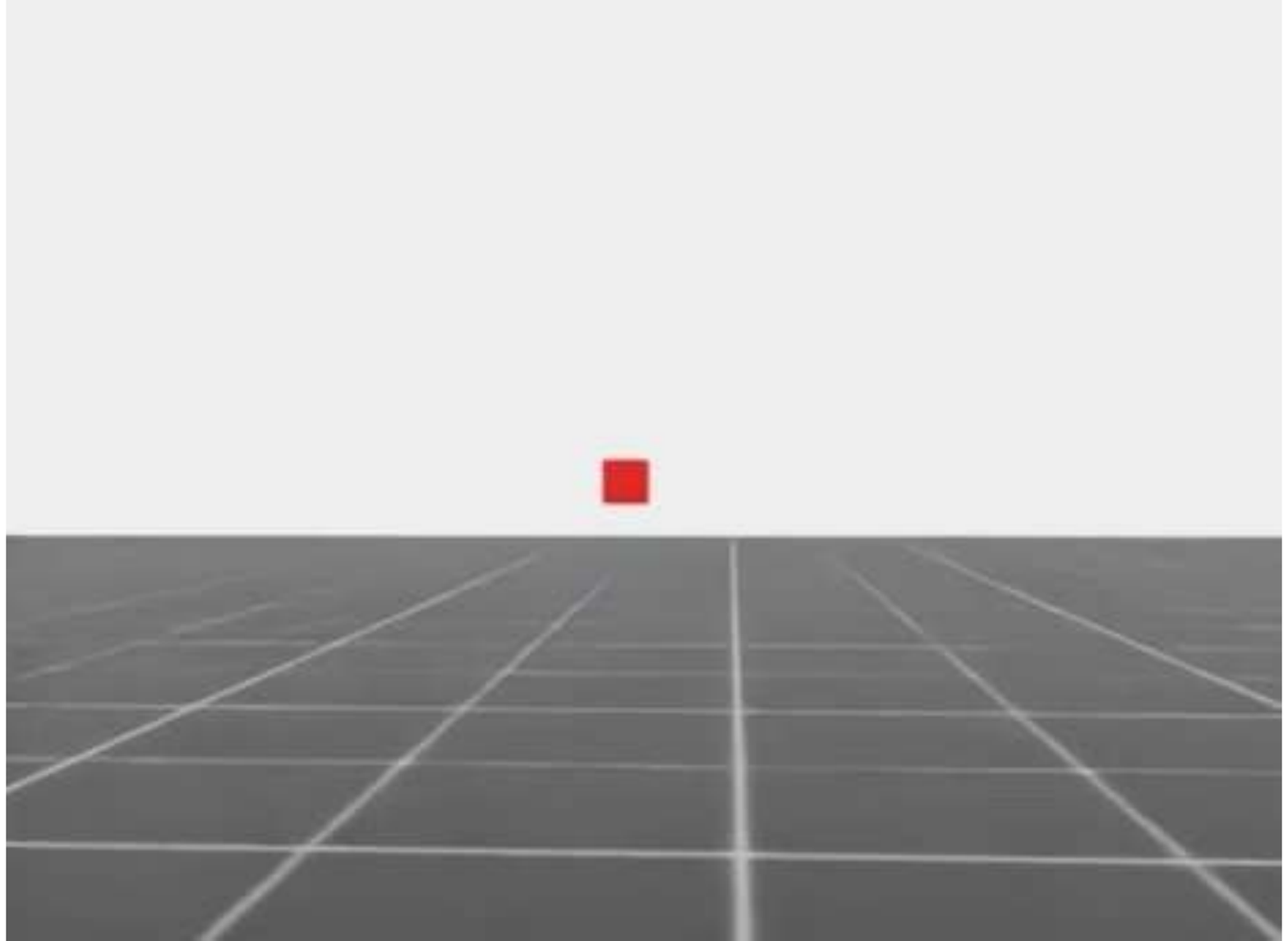}
\label{fig:dataset_single}}
\hfill
\subfloat[Multi-color heterogeneous scene (Dataset~2).]{
\includegraphics[width=0.3\linewidth,
trim=0 0cm 0 0,clip]{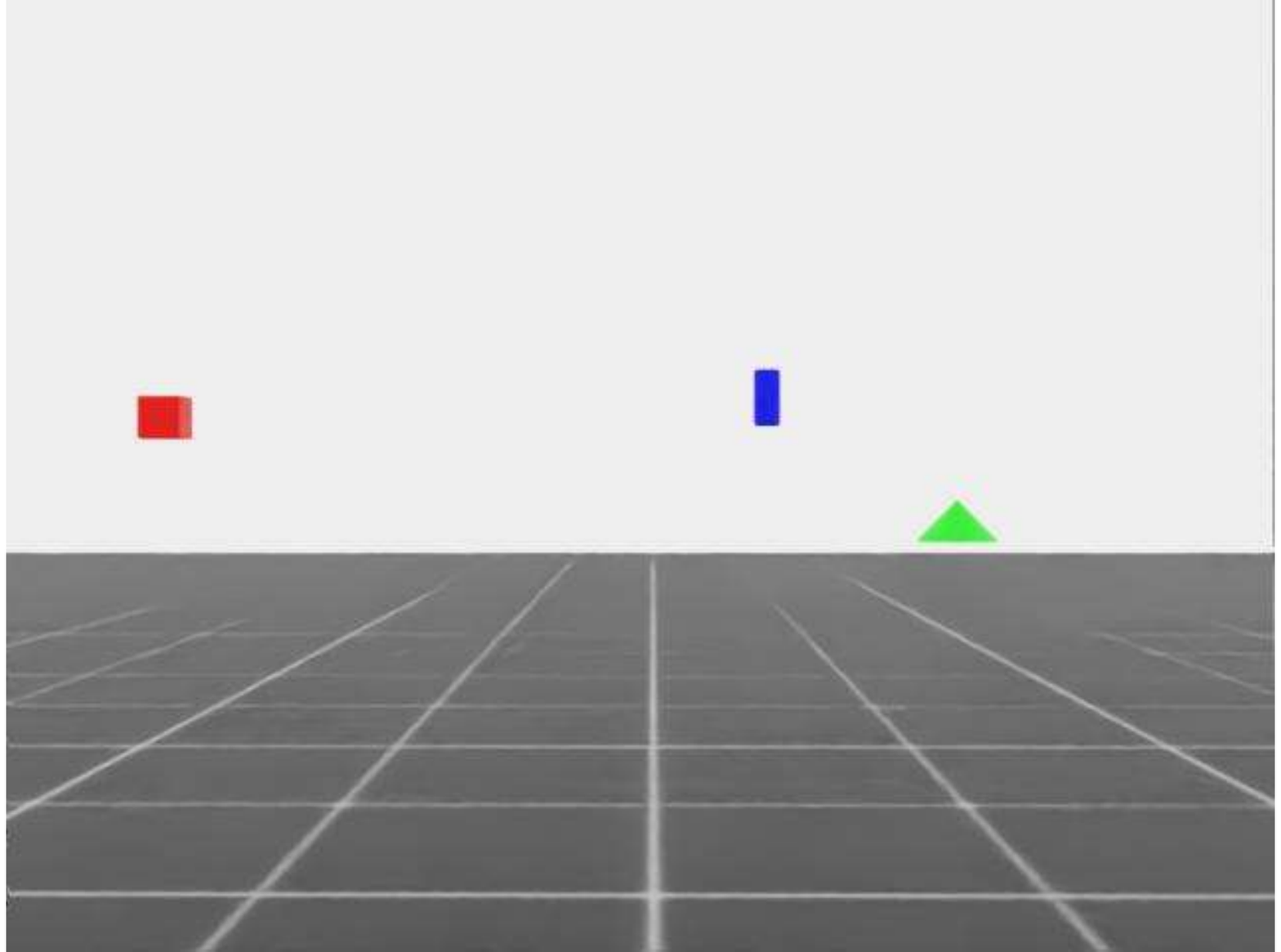}
\label{fig:dataset_multi}}
\hfill
\subfloat[An example of drone's trajectories.]{
\includegraphics[width=0.3\linewidth,
trim=1cm 0cm 0cm 0cm,clip]{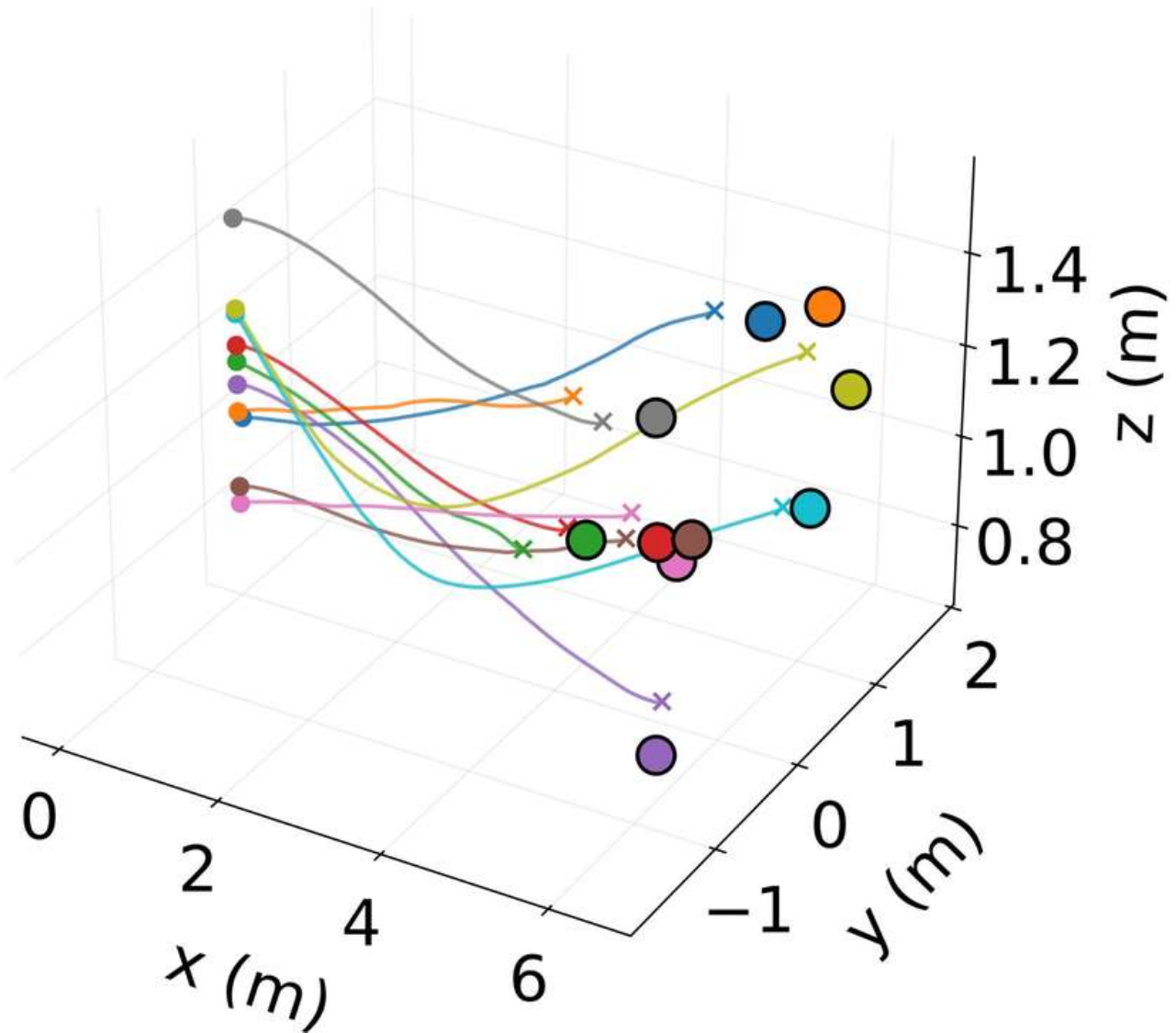}
\label{fig:traj3d}}

\subfloat[The projections onto the $\mathrm{xy}$ plane.]{
\includegraphics[width=0.3\linewidth,
trim=0cm 0cm 0 0,clip]{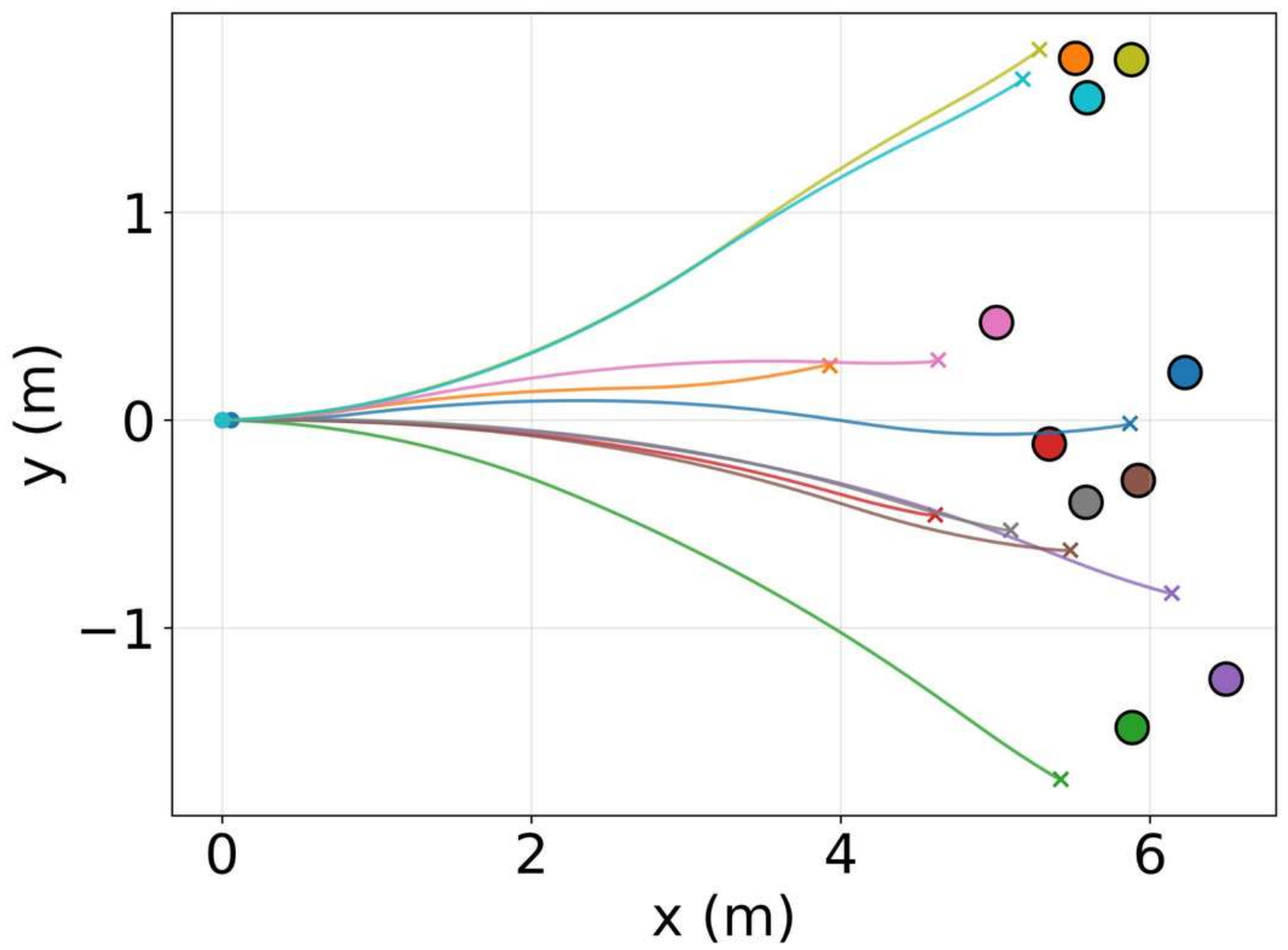}
\label{fig:traj_xy}}
\hfill
\subfloat[The projections onto the $\mathrm{xz}$ plane.]{
\includegraphics[width=0.3\linewidth,
trim=0cm 0cm 0 0,clip]{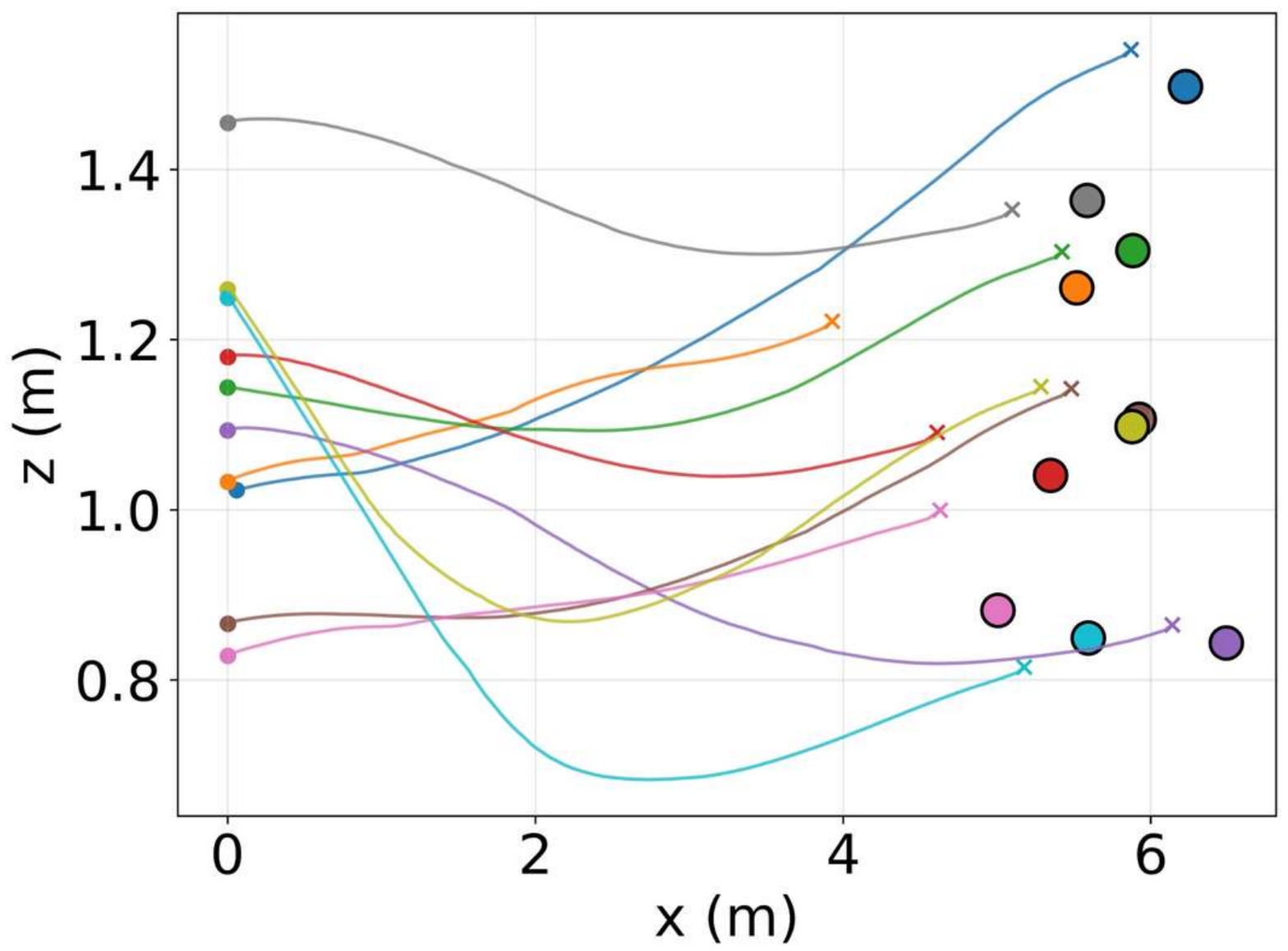}
\label{fig:traj_xz}}
\hfill
\subfloat[The projections onto the $\mathrm{yz}$ plane.]{
\includegraphics[width=0.3\linewidth,
trim=0cm 0cm 0 0,clip]{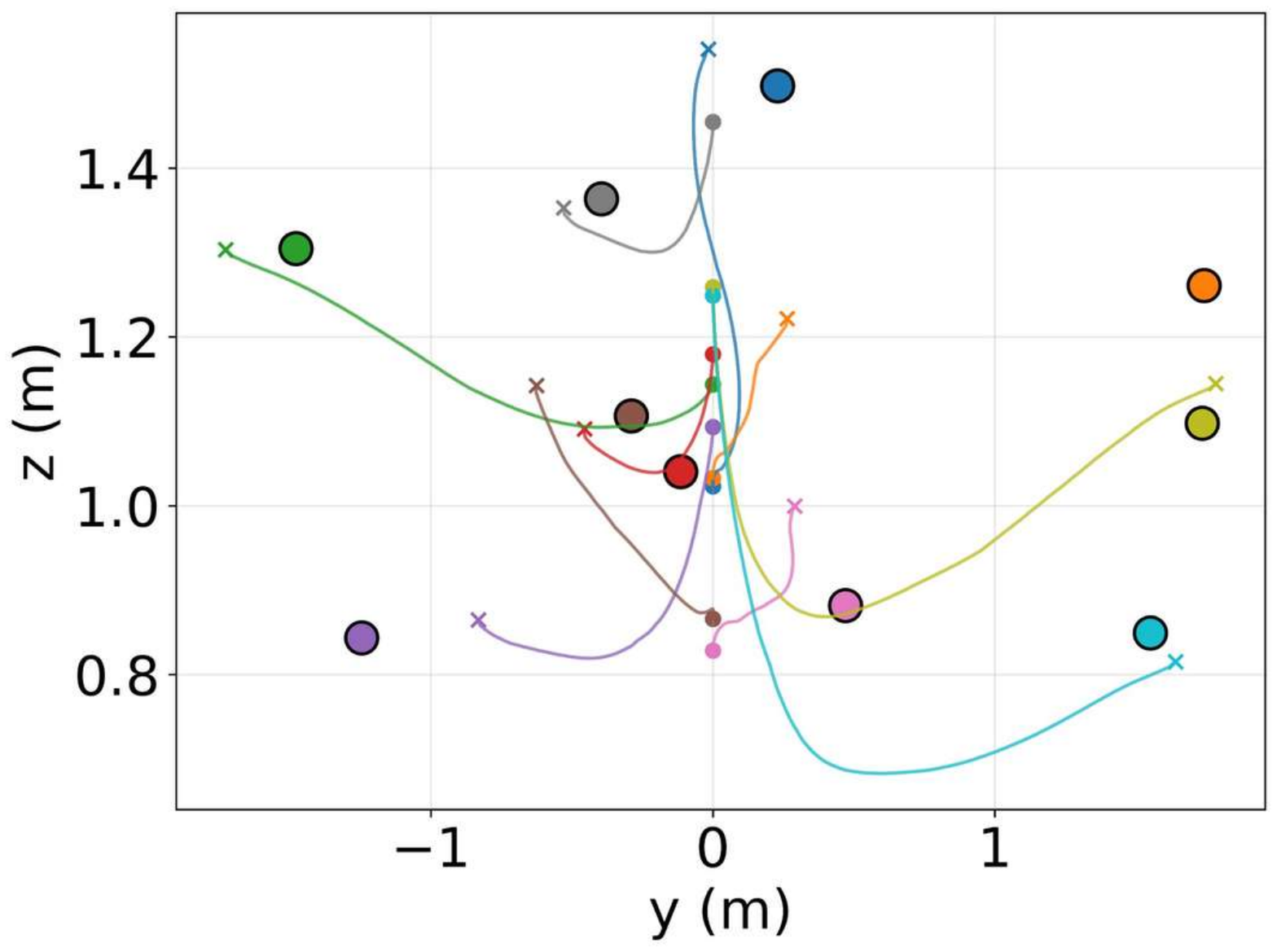}
\label{fig:traj_yz}}

\caption{Training environments: \protect\subref{fig:dataset_single} single-object dataset (Dataset~1) and \protect\subref{fig:dataset_multi} multi-object RGB-shapes dataset with targets varying in color and shape (Dataset~2). \protect\subref{fig:traj3d} Example drone trajectories for \texttt{Fly to the red cube} over 10 episodes using $\pi_{0.5}$ backbone. The 3D trajectory is shown with projections onto the $\mathrm{xy}$ \protect\subref{fig:traj_xy}, $\mathrm{xz}$ \protect\subref{fig:traj_xz}, and $\mathrm{yz}$ \protect\subref{fig:traj_yz} planes. Dots mark start positions, crosses denote final positions, and colored circles indicate target objects.}
\label{fig:simeval}

\end{figure}

To clearly distinguish data collection environments from evaluation conditions, we collect training data in two environments and introduce a third environment exclusively for evaluation. Specifically, we first construct (i) a canonical single-object environment containing only a red cube, which serves as the baseline data collection setting. We then build (ii) a multi-object heterogeneous environment containing three target objects with distinct color-shape combinations: a red cube, a green cone, and a blue cylinder. In addition, to assess generalization beyond the training configurations, we introduce (iii) a color-permuted heterogeneous environment used only for evaluation, in which the same object categories are retained but their color assignments are reassigned to form a green cube, a red cylinder, and a blue cone. Representative onboard camera observations from the two data collection environments are shown in Fig.~\ref{fig:dataset_single} and Fig.~\ref{fig:dataset_multi}, respectively.

\text{(i) Canonical single-object baseline:} A single red cube is placed at
randomly sampled positions within a bounded region. The language instruction is
fixed to \texttt{Fly to the red cube.}

\text{(ii) Multi-color heterogeneous scene:} Three objects of varying colors and
shapes (red cube, green cone, and blue cylinder) are present simultaneously. The
language instruction specifies both color and shape, e.g., \texttt{Fly to the
green cone.}

\text{(iii) Color-permuted heterogeneous scene:} The same three objects are present but the colors are permuted. (green cube, red cylinder, and blue cone). The language instruction again specifies both color and shape, e.g., \texttt{Fly to the green cube.}

\subsection{Results and discussion}
Tables~\ref{tab:single-objectsetting}, \ref{tab:multi-object-setting}, and \ref{tab:multinovelobject} report the success rates of our proposed pipeline using the $\pi_{0.5}$ and SmolVLA backbones, denoted as \ac{Exp2VLA}($\pi_{0.5}$) and \ac{Exp2VLA}({\footnotesize SmolVLA}), respectively. We evaluate both variants in Isaac Lab under closed-loop control across three predefined scenarios: (i) the canonical single-object baseline, (ii) the multi-object heterogeneous scene, and (iii) the color-permuted heterogeneous evaluation scene.

When the evaluation scenarios use the same object--color--shape configurations present in the training dataset (see Fig.~\ref{fig:dataset_multi}), the model consistently demonstrates high performance, achieving a success rate of 68.0\% for the red cube, 64.4\% for the blue cylinder, and 62.6\% for the green cone. The drone's trajectories for the trained-configuration evaluation are visualized in Fig.~\ref{fig:traj3d}-\ref{fig:traj_yz}, which illustrates the drone's world-frame convergence across 10 evaluation episodes. To assess zero-shot generalization, an additional evaluation was conducted using the color-permuted heterogeneous scene (iii), which contains previously unseen object--color configurations: the object colors were permuted, and the novel command \texttt{Fly to the green cube}, \texttt{Fly to the red cylinder}, and \texttt{Fly to the blue cone} were issued. Although these specific combinations were never encountered during training, \ac{Exp2VLA}($\pi_{0.5}$) maintained an average 51.7\% success rate (see Table~\ref{tab:multinovelobject}), demonstrating robust semantic generalization and the ability to map learned spatial features to novel object--color pairings.

In contrast, comparative evaluations with the \ac{Exp2VLA}({\footnotesize SmolVLA}) highlight the challenges of deploying ultra-compact \ac{vla}s for high-frequency drone control. While the \ac{Exp2VLA}({\footnotesize SmolVLA}) achieved a 46.6\% success rate in Canonical single-object environment (see Fig.~\ref{fig:dataset_single} and Table~\ref{tab:single-objectsetting}) using configurations seen during training, its performance degraded significantly to 15\% in the multi-object RGB-shapes environment (see Fig.~\ref{fig:dataset_multi} and Table~\ref{tab:multi-object-setting}). This suggests that while such models are efficient, they struggle with the complex spatial-temporal demands and long-horizon requirements of aerial navigation.

\begin{figure}[t!]
\centering
\vspace{3pt}
\begin{tikzpicture}
\begin{axis}[
    ybar,
    width=\linewidth,
    height=3.8cm,
    bar width=17pt,
    ymin=0,
    ymax=50,
    ylabel={Loop latency (ms)},
    symbolic x coords={Pi05,SmolVLA},
    xtick=data,
    xticklabels={{$\pi_{0.5}$},{SmolVLA}},
    enlarge x limits=0.55,
    ymajorgrids=true,
    grid style={gray!20},
    nodes near coords,
    nodes near coords align={vertical},
    every node near coord/.append style={font=\scriptsize},
    tick label style={font=\scriptsize},
    label style={font=\scriptsize},
]
\addplot+[
    draw=blue!70!black,
    fill=blue!30,
    error bars/y dir=both,
    error bars/y explicit,
]
    coordinates {(Pi05,26.53) +- (0,16.68) (SmolVLA,23.22) +- (0,13.73)};
\end{axis}
\end{tikzpicture}
\caption{Measured inference-loop latency for \ac{Exp2VLA}($\pi_{0.5}$) and \ac{Exp2VLA}({\footnotesize SmolVLA}). Bars show mean loop time with error bars indicating one standard deviation. \ac{Exp2VLA}($\pi_{0.5}$) requires 12.90\,GB VRAM, while the compact \ac{Exp2VLA}({\footnotesize SmolVLA}) requires only 6.739\,GB.}
\label{fig:inference_latency}
\end{figure}
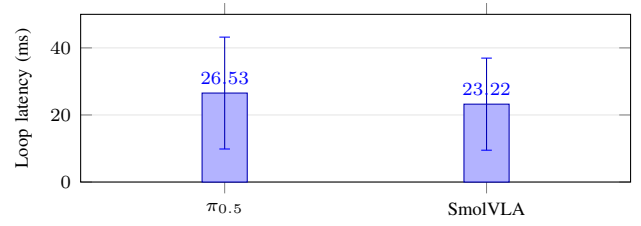

To complement the task-success evaluation, we also profile the closed-loop inference time and memory footprint of both backbones. Fig.~\ref{fig:inference_latency} summarizes the measured inference-loop latency along with the total VRAM usage for each backbone. Although \ac{Exp2VLA}({\footnotesize SmolVLA}) reduces the total VRAM requirement from 12.90~GB to 6.739~GB, its average loop time is only slightly lower than that of \ac{Exp2VLA}($\pi_{0.5}$). Thus, the practical trade-off is primarily between memory footprint and navigation performance rather than a large latency difference.

\begin{table}[htbp!]
  \centering
  \caption{Canonical single-object setting: 500 episodes per task, acceptance radius $\epsilon = 0.60$\,m, and target object includes a single red object only.}
  \label{tab:single-objectsetting}
  \renewcommand{\arraystretch}{0.93}
  \setlength{\tabcolsep}{3pt}
  \scriptsize
  \begin{tabular}{
      >{\raggedright\arraybackslash}p{1.4cm}
      >{\raggedright\arraybackslash}p{2.2cm}
      >{\raggedright\arraybackslash}p{2.2cm}
      }
    \toprule
    \textbf{Backbone} &
    \textbf{Target object} &
    \textbf{Success rate} (\%)\\
    \midrule
    \shortstack{\ac{Exp2VLA}\\($\pi_{0.5}$)}
      & Red cube & 84.10\\
    \midrule
    \shortstack{\ac{Exp2VLA}\\({\footnotesize SmolVLA})}  & Red cube & 46.60 \\
    \bottomrule
  \end{tabular}
\end{table}

\begin{table}[htbp!]
  \centering
  \caption{Multi-color heterogeneous scene: 500 episodes per task with an acceptance radius of $\epsilon = 0.60$\,m. Target objects include a red cube, blue cylinder, and green cone. $\pi_{0.5}$ and SmolVLA contain 4B and 0.45B parameters, respectively.}
  \label{tab:multi-object-setting}
  \renewcommand{\arraystretch}{0.93}
  \setlength{\tabcolsep}{3pt}
  \scriptsize

  \begin{tabular}{
      >{\raggedright\arraybackslash}p{1.2cm}
      >{\raggedright\arraybackslash}p{1.9cm}
      >{\raggedright\arraybackslash}p{1.15cm}
      >{\raggedright\arraybackslash}p{1.55cm}
      }
    \toprule
    \textbf{Backbone} &
    \textbf{Target object} &
    \textbf{Success rate} (\%) &
    \textbf{Avg. success rate} (\%)\\
    \midrule

    \multirow{3}{*}{\shortstack{\ac{Exp2VLA}\\($\pi_{0.5}$)}}
        & Red cube      & 68.00 &  \\
        & Blue cylinder & 64.40 & 65.00 \\
        & Green cone    & 62.60 &       \\
    \midrule
    \multirow{3}{*}{\shortstack{\ac{Exp2VLA}\\({\footnotesize SmolVLA})}} & Red cube      & 16.00 &  \\
        & Blue cylinder & 14.00 &  15.00    \\
        & Green cone    & 15.00 &       \\
    \bottomrule
  \end{tabular}
\end{table}

\begin{table}[htbp!]
  \centering
  \caption{Color-Permuted Heterogeneous scene setting: 500 episodes per task, acceptance radius $\epsilon = 0.60$\,m, and target object includes unseen objects. $^\dagger$ Zero-shot evaluation on unseen objects in training.}
  \label{tab:multinovelobject}
  \renewcommand{\arraystretch}{0.93}
  \setlength{\tabcolsep}{3pt}
  \scriptsize

  \begin{tabular}{
      >{\raggedright\arraybackslash}p{1.2cm}
      >{\raggedright\arraybackslash}p{1.9cm}
      >{\raggedright\arraybackslash}p{1.15cm}
      >{\raggedright\arraybackslash}p{1.55cm}
      }
    \toprule
    \textbf{Backbone} &
    \textbf{Target objects} &
    \textbf{Success rate} (\%) &
    \textbf{Avg. success rate} (\%)\\
    \midrule

    \multirow{3}{*}{\shortstack{\ac{Exp2VLA}\\($\pi_{0.5}$)}}
        & Green cube$^\dagger$      & 51.0 &  \\
        & Red cylinder$^\dagger$  & 75.0 &   51.7    \\
        & Blue cone$^\dagger$    & 29.0 &       \\
    \bottomrule
  \end{tabular}
\end{table}

Analysis of the results reveals that the performance disparity between $\pi_{0.5}$ and SmolVLA underscores the critical role of visual encoder capacity in enabling effective multi-object aerial navigation. The PaliGemma (3B) backbone in $\pi_{0.5}$ provides sufficiently rich visual--semantic features to distinguish between multiple objects of different colors and shapes, whereas the compact SmolVLA encoder appears unable to maintain discriminative representations when visual distractors are present. Second, the moderate drop in success rate from trained configurations ($\sim$65\%) to the previously unseen configuration (51.7\%) suggests that while $\pi_{0.5}$ has acquired compositional color--shape grounding, this capability remains imperfect—the model can generalize to novel pairings, but not with the same reliability as for configurations encountered during training. Third, the consistent performance across three distinct target objects (red cube, blue cylinder, green cone) indicates that the model does not overfit to a single object type, supporting the effectiveness of the multi-task training strategy with diverse language instructions. Overall, the results indicate that the \ac{Exp2VLA}($\pi_{0.5}$) architecture provides a favorable balance of reasoning capacity and inference speed, effectively adapting to complex task environments where standard compact \ac{vla}s fail to generalize.

\subsection{Simulation in the loop}
\begin{figure}[!b]
\centering
\subfloat[The SITL pipeline.]{
\includegraphics[width=0.85\linewidth, trim=0cm 0cm 0 0, clip]{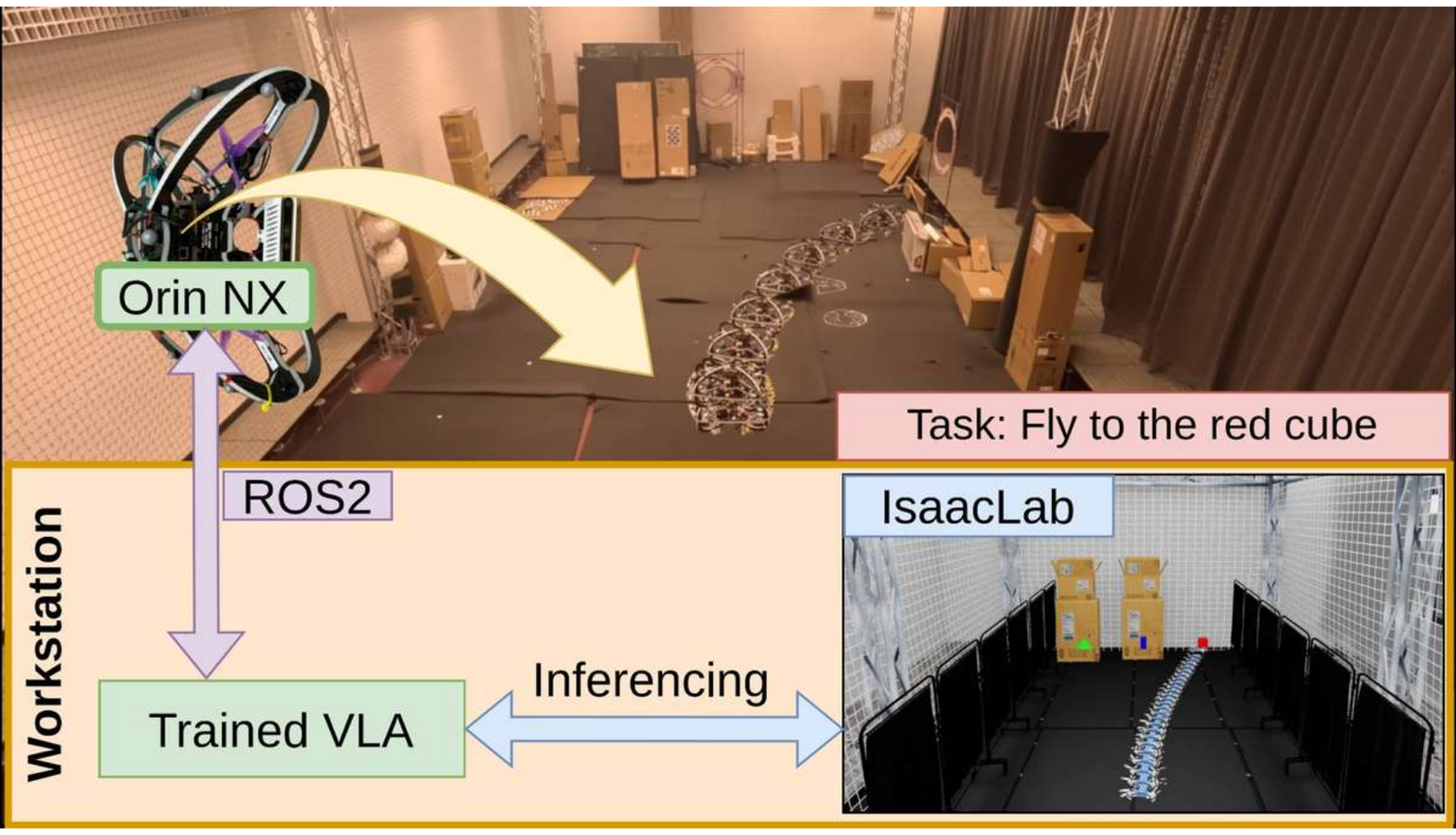}
\label{fig:sitl_arc}}
\\[0.5em]

\subfloat[Task: \texttt{Fly to the green cone}]{
\includegraphics[width=0.3\linewidth, trim=0cm 0cm 0 0, clip]{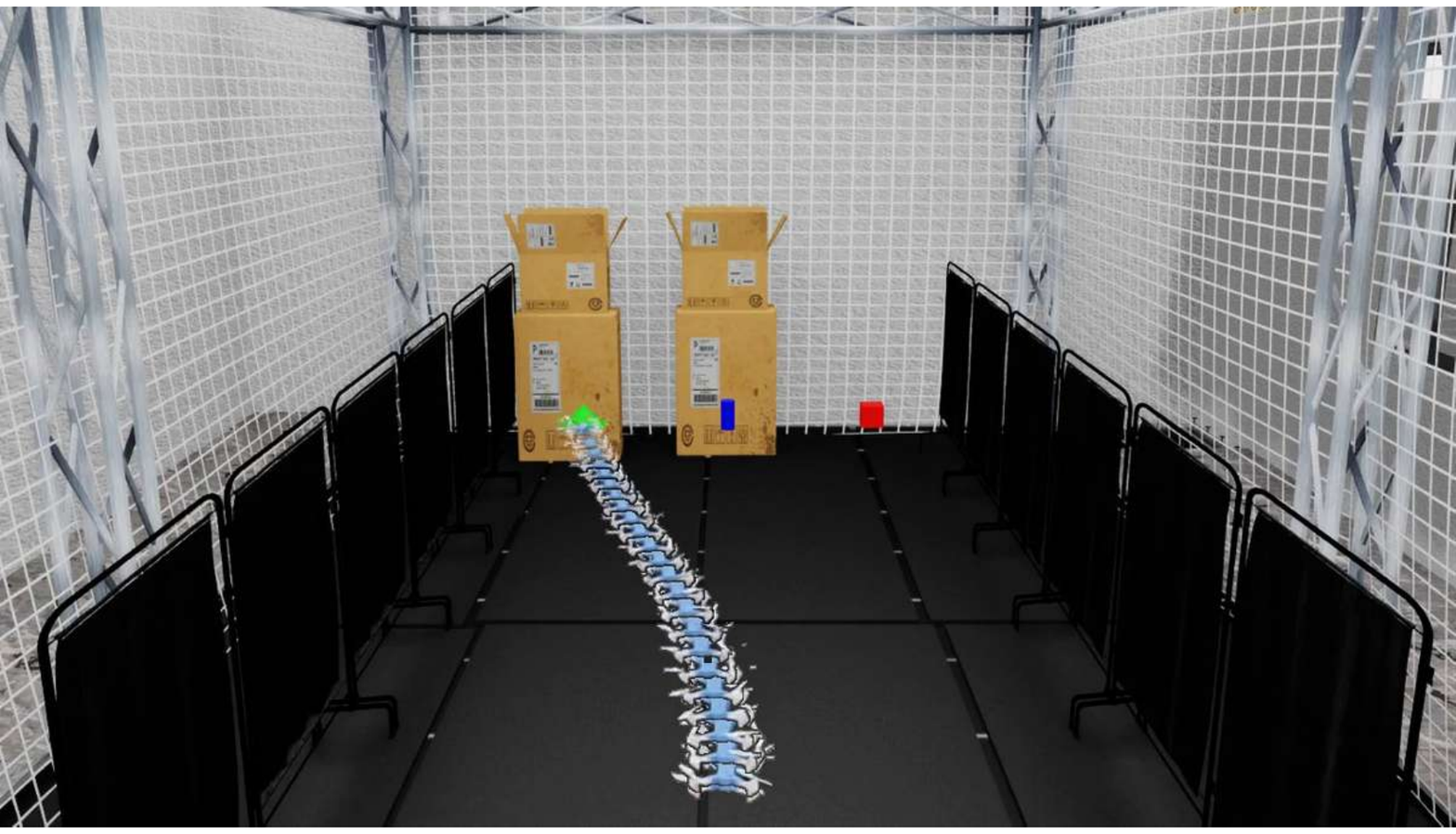}
\label{fig:fly2greencone}}
\hfill
\subfloat[Task: \texttt{Fly to the blue cylinder}]{
\includegraphics[width=0.3\linewidth, trim=0cm 0cm 0 0, clip]{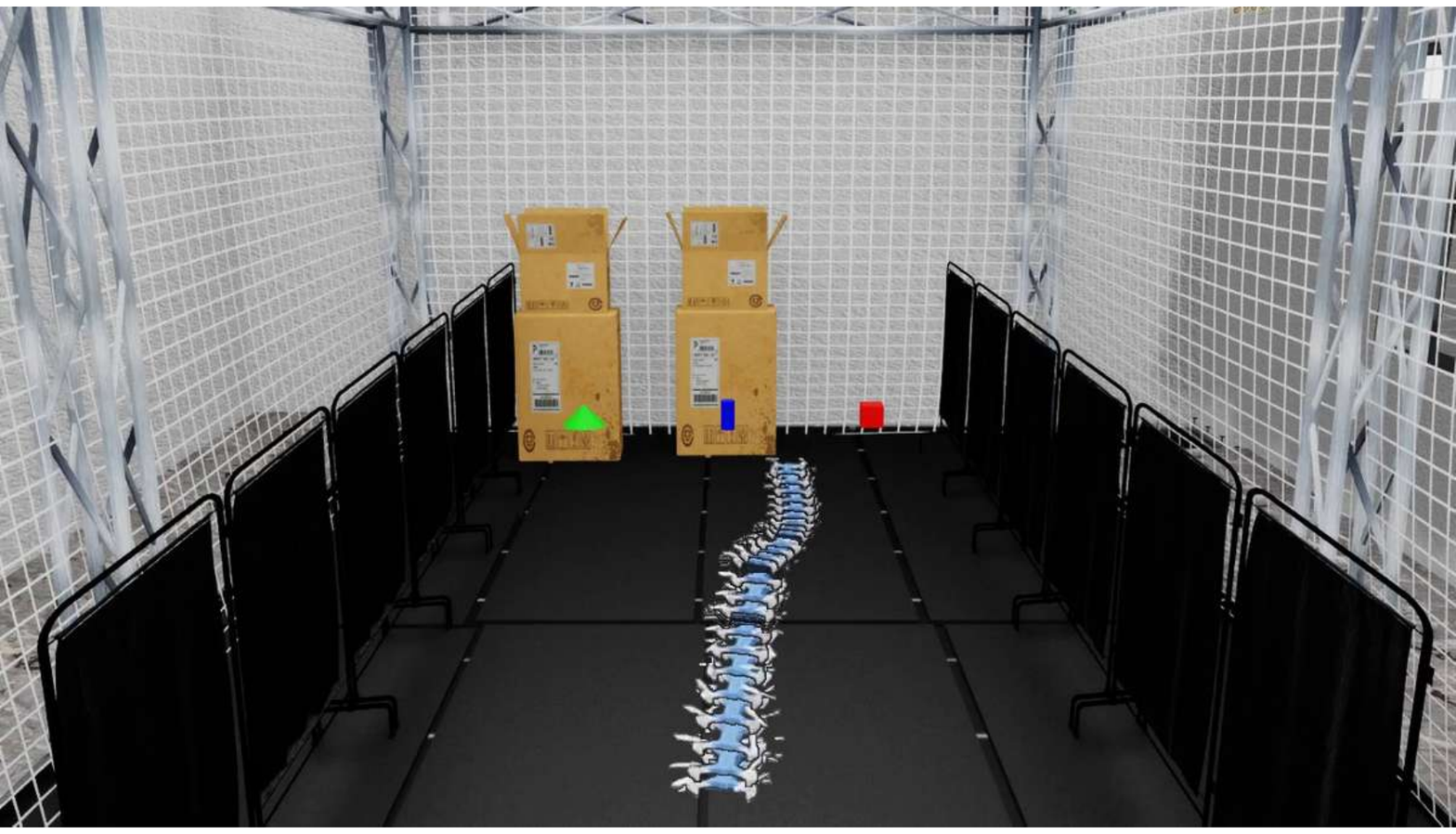}
\label{fig:fly2cylinder}}
\hfill
\subfloat[Task: \texttt{Fly to the red cube}]{
\includegraphics[width=0.3\linewidth, trim=0cm 0cm 0cm 0cm, clip]{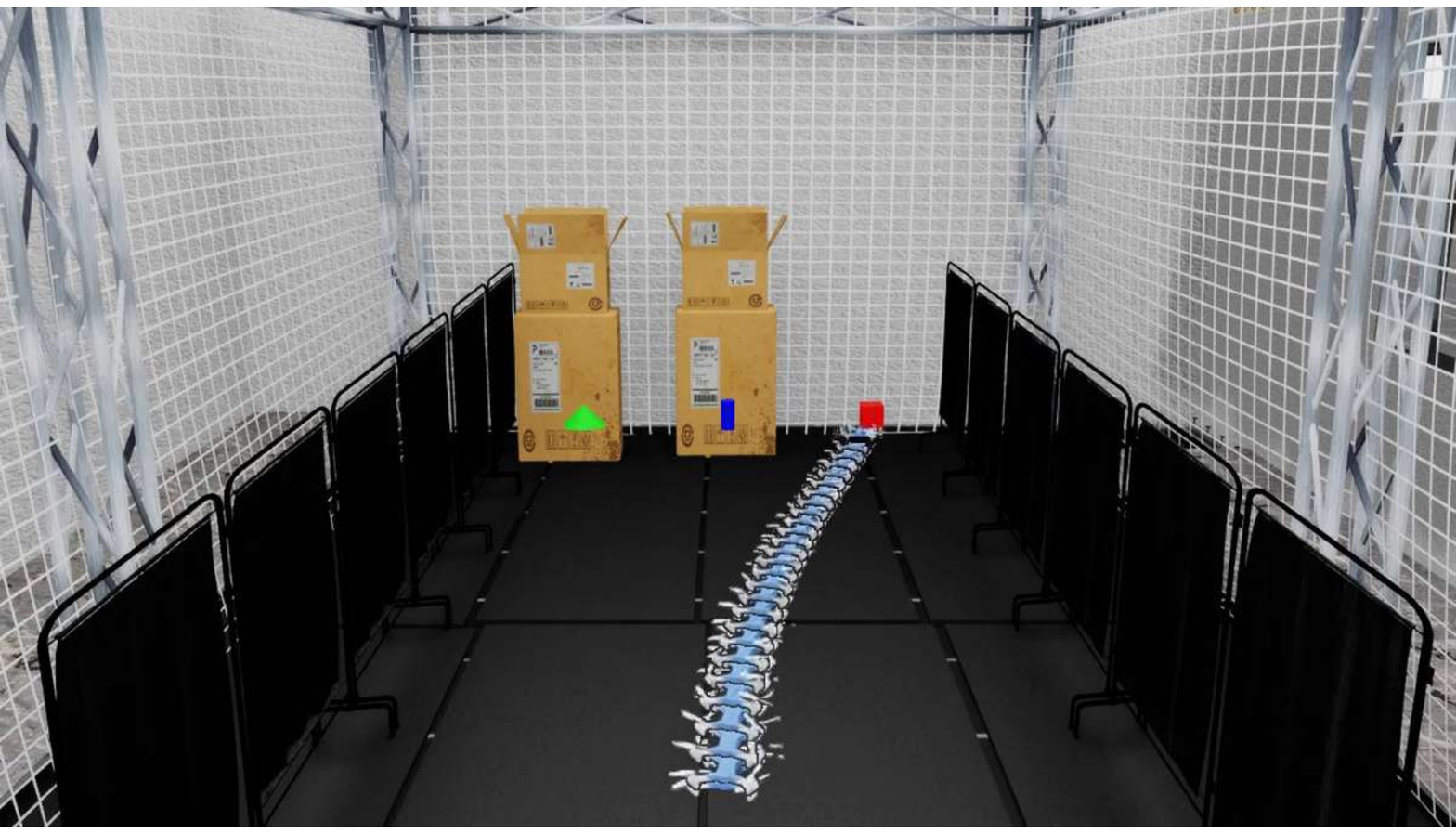}
\label{fig:fly2redcube}}
\\

\subfloat[Task: \texttt{Fly to the green cone}]{
\includegraphics[width=0.3\linewidth, trim=0cm 0cm 0 0, clip]{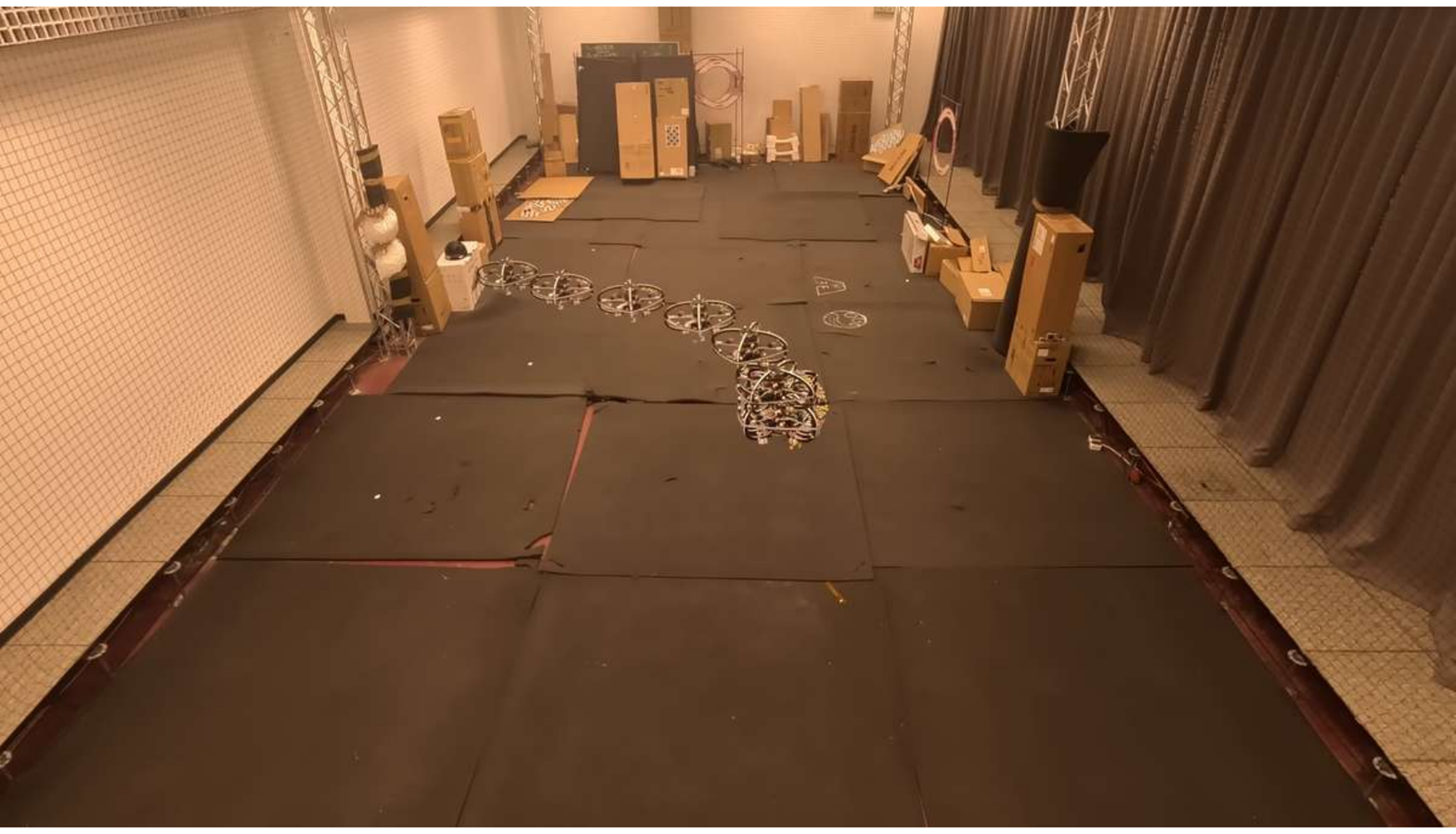}
\label{fig:sitl_fly2greencone}}
\hfill
\subfloat[Task: \texttt{Fly to the blue cylinder}]{
\includegraphics[width=0.3\linewidth, trim=0cm 0cm 0 0, clip]{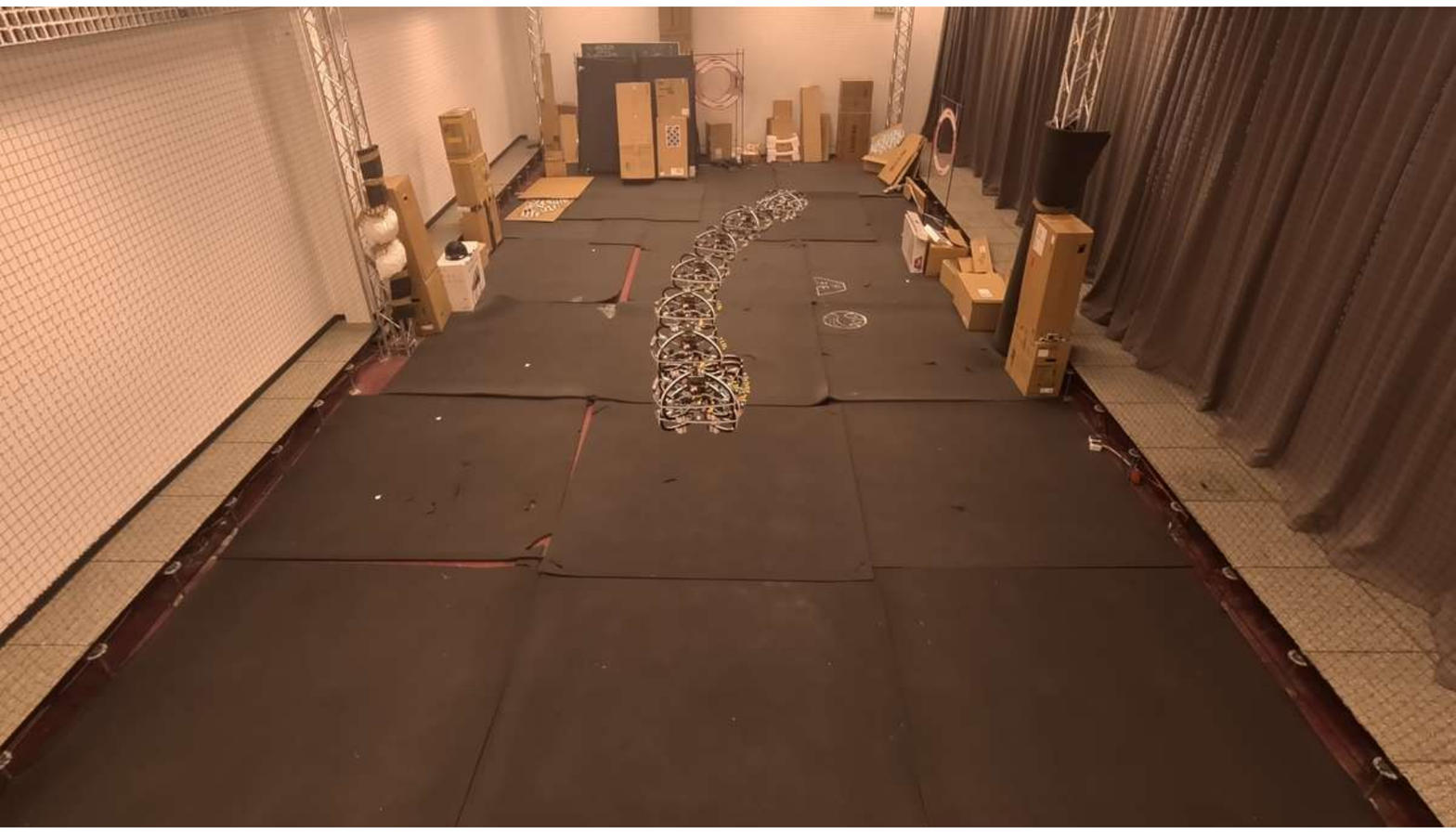}
\label{fig:sitl_fly2cylinder}}
\hfill
\subfloat[Task: \texttt{Fly to the red cube}]{
\includegraphics[width=0.3\linewidth, trim=0cm 0cm 0cm 0cm, clip]{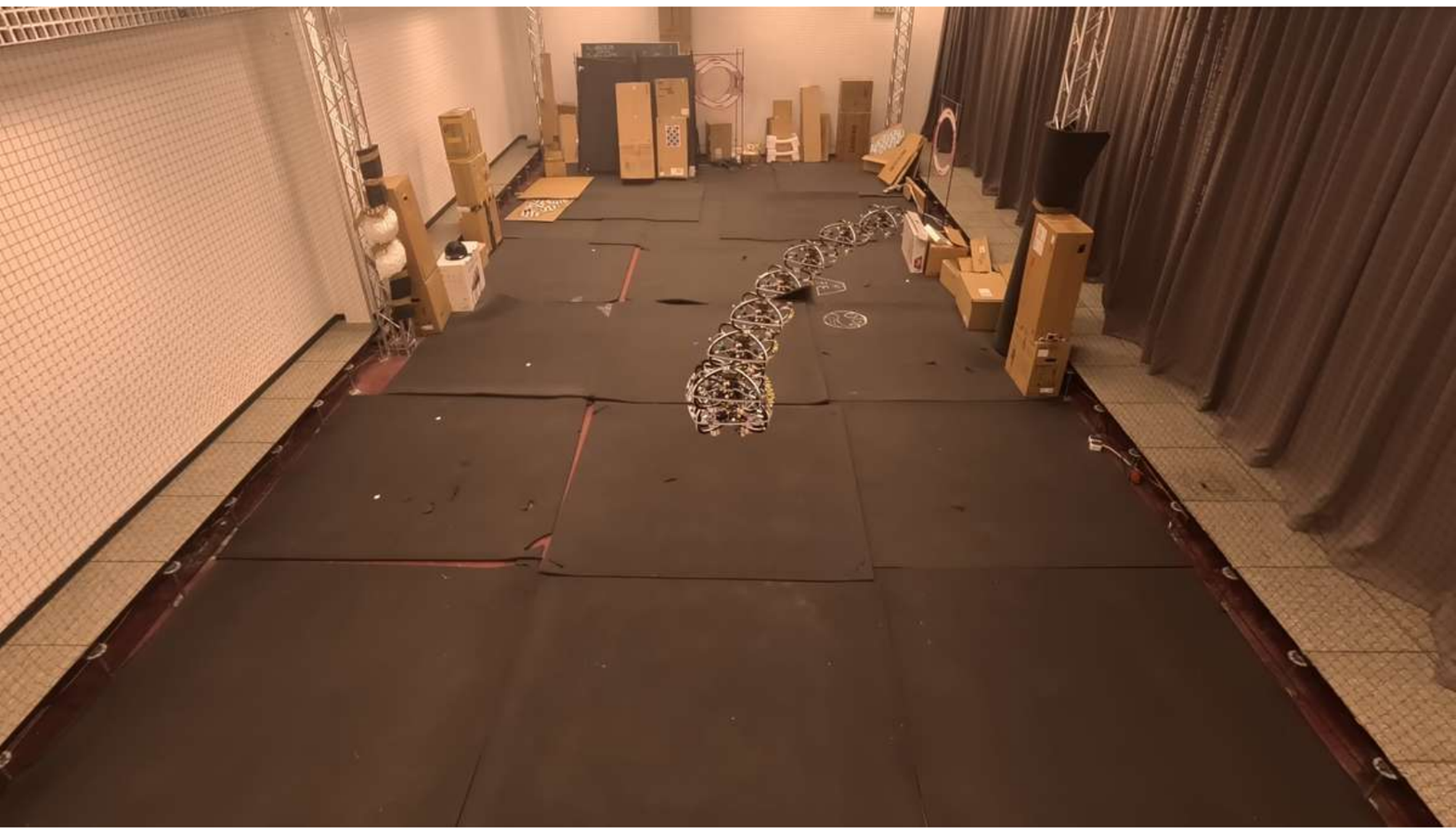}
\label{fig:sitl_fly2redcube}}

\caption{The top row (a) presents the SITL pipeline, in which the trained VLA performs inference on a workstation using Isaac Lab and sends commands to the real drone via ROS2. The middle row (b-d) shows sim-to-sim results on three target-reaching tasks, and the bottom row (e-g) demonstrates corresponding real-time executions on the physical platform, highlighting the feasibility of deploying the fine-tuned VLA model in a real-time SITL setting.}
\label{fig:sitl}
\end{figure}

To demonstrate both generalizability and real-time performance, we first conduct sim-to-sim and simulation-in-the-loop (SITL) experiments. Fig.~\ref{fig:sitl_arc} illustrates the overall SITL pipeline. 
In the sim-to-sim setup, we build a simulation environment that matches our lab at a one-to-one scale and perform inference on three representative tasks: navigating to a green cone, a blue cylinder, and a red cube, as shown in Fig.~\ref{fig:fly2greencone}-\ref{fig:fly2redcube}.
We then evaluate real-time deployment in an SITL setup, where the trained \ac{Exp2VLA}($\pi_{0.5}$) is executed on a mobile workstation, since the onboard Orin NX lacks sufficient computational resources to run the full VLA model. The generated control commands are transmitted via ROS2 to the drone in real time. The results in Fig.~\ref{fig:sitl_fly2greencone}-\ref{fig:sitl_fly2redcube} demonstrate the feasibility of deploying the trained VLA model under this off-board inference configuration.
\section{Conclusion and Future Work}\label{sec:conclusion}
Deploying \ac{vla} models on aerial robots remains challenging due to the computational constraints of \ac{uav} platforms and the scarcity of domain-specific training data. In this work, we present a lightweight end-to-end pipeline for language-conditioned drone navigation built upon the $\pi_{0.5}$ architecture. Our framework integrates expert-driven data collection in simulation, automated dataset conversion to the LeRobot format, efficient fine-tuning on a single consumer-grade GPU with a frozen vision encoder, and closed-loop evaluation in Isaac Lab. Experimental results show that \ac{Exp2VLA}($\pi_{0.5}$) achieves up to 68\% task success on in-distribution scenarios and maintains 51.7\% on zero-shot out-of-distribution configurations involving previously unseen object--color pairings. In contrast, the ultra-compact \ac{Exp2VLA}({\footnotesize SmolVLA}) drops to 15\% in multi-object settings, highlighting that sufficient visual encoder capacity is essential for discriminating among multiple targets in complex aerial scenes. The consistent performance across diverse target objects further confirms the effectiveness of multi-task training with varied language instructions. These findings suggest that mid-scale \ac{vla} architectures can provide a practical trade-off between reasoning capability and deployment efficiency for resource-constrained aerial platforms.

Future work will improve these models by examining the semantic depth of the action encoders in the interpretation of abstract linguistic cues beyond zero-shot tasks. Additionally, it is essential to address the lack of a standardized common environment for benchmarking drone VLA policies.
\vspace{-0.4em}
\section*{Acknowledgment}
The authors gratefully acknowledge the funding of this project by computing time provided by the Paderborn Center for Parallel Computing (PC\textsuperscript{2}). This work was partially supported by the Horizon Europe Grant Agreements No.~101136056 and No.~101119774.
\vspace{-0.5em}
\bibliographystyle{IEEEtran}
\bibliography{references}

\end{document}